\documentclass[11pt]{article}
\usepackage{latex/acl}
\usepackage{times}
\usepackage{latexsym}
\usepackage[T1]{fontenc}
\usepackage[utf8]{inputenc}
\usepackage{microtype}
\usepackage{inconsolata}
\usepackage{graphicx}
\usepackage{amsmath}
\usepackage{amssymb}
\usepackage{booktabs}
\usepackage{multirow}
\usepackage{xcolor}
\usepackage{algorithm}
\usepackage{algorithmic}
\usepackage{subcaption}

\title{SPD-RAG: Sub-Agent Per Document Retrieval-Augmented Generation}

\author{
  Yagiz Can Akay\textsuperscript{1} \quad Muhammed Yusuf Kartal\textsuperscript{1} \quad Esra Alparslan\textsuperscript{1}  \\ \textbf{Faruk Ortakoyluoglu\textsuperscript{1}}
  \quad \textbf{Arda Akpinar\textsuperscript{2}} \\
  \textsuperscript{1}TOBB University of Economics and Technology \quad \textsuperscript{2}OSTIM Technical University \\
  \texttt{\{y.akay, m.kartal, ealparslan, fortakoyluoglu\}@etu.edu.tr}
  \texttt{\ 230206003@ostimteknik.edu.tr}
}

\begin{document}
\maketitle

\begin{abstract}
Answering complex, real-world queries often requires synthesizing facts scattered across vast document corpora. In these settings, standard retrieval-augmented generation (RAG) pipelines suffer from incomplete evidence coverage, while long-context large language models (LLMs) struggle to reason reliably over massive inputs. We introduce SPD-RAG, a hierarchical multi-agent framework for exhaustive cross-document question answering that decomposes the problem along the document axis. Each document is processed by a dedicated document-level agent operating only on its own content, enabling focused retrieval, while a coordinator dispatches tasks to relevant agents and aggregates their partial answers. Agent outputs are synthesized by merging partial answers through a token-bounded synthesis layer (which supports recursive map-reduce for massive corpora). This document-level specialization with centralized fusion improves scalability and answer quality in heterogeneous multi-document settings while yielding a modular, extensible retrieval pipeline. On the LOONG benchmark (EMNLP 2024) for long-context multi-document QA, SPD-RAG achieves an Avg Score of 58.1 (GPT-5 evaluation), outperforming Normal RAG (33.0) and Agentic RAG (32.8) while using only 38\% of the API cost of a full-context baseline (68.0).
\end{abstract}

\section{Introduction}
\label{sec:introduction}

Large language models (LLMs) and the agentic systems built around them are increasingly used for complex information search tasks~\cite{guo2024large}. Real-world questions often require synthesizing evidence scattered across many documents, such as assessing a company's financial risks across years of reports or integrating findings from multiple scientific papers.\footnote{Examples inspired by scenarios in the Loong benchmark~\cite{wang2024leave}.} Such questions are natural and common, yet expose critical coverage and reasoning bottlenecks in current NLP systems.

Traditional single-index or single-agent retrieval systems struggle to maintain both scalability and contextual relevance over large, heterogeneous corpora, especially for long or structurally complex documents. Standard RAG pipelines retrieve a fixed number of documents $K$ and process them within a single context window~\cite{lewis2020retrieval}, which fails when answers depend on information distributed across many documents, since evidence beyond the top-$K$ results is typically discarded in a single retrieval pass. Long-context LLMs extend context windows to 128K--2M tokens~\cite{team2024gemini}, but empirical evidence shows that reasoning quality degrades significantly as context length increases~\cite{liu2023lost}. This suggests that the bottleneck is not only retrieval, but also \textit{reasoning at scale} over many documents and hundreds of thousands of tokens.

In this work, we introduce \textsc{SPD-RAG} (Sub-agent per Document Retrieval-Augmented Generation), a hierarchical multi-agent architecture for \textit{exhaustive} multi-document question answering that factors the problem along the document axis rather than the task axis. Instead of forcing a single model to hunt through a massive global index, \textsc{SPD-RAG} uses a central coordinator to decompose the user's query into shared instructions. It then deploys a dedicated, cost-efficient sub-agent to \textit{each} document in the corpus. These agents operate entirely in parallel, treating their assigned documents as isolated retrieval universes to extract relevant findings. Finally, a synthesis model aggregates these document-grounded findings---employing a recursive merging fallback for exceptionally large corpora---to construct a comprehensive, final answer. Our codebase and datasets used in this paper can be found at \\
\url{https://github.com/NebulAICompany/SPD-RAG}.

Our contributions are as follows:
\begin{itemize}
    \item We propose \textsc{SPD-RAG}, a hierarchical multi-agent framework that combines per-document agentic RAG with a centralized synthesis layer, where cost-efficient document agents are coordinated by a smarter coordinator agent, enabling document-level specialization and parallel execution while allowing each document to be analyzed in depth to ensure that all relevant information is incorporated without missing critical evidence.
    \item We evaluate on the Loong benchmark~\cite{wang2024leave}, which includes multi-document QA instances over financial reports and academic papers with an average of 11 documents per instance and context lengths from 10K to beyond 250K tokens, and show that our system substantially outperforms standard RAG and Agentic RAG baselines under GPT-5-judged Avg Score (adapted from Loong's GPT-4-judged protocol; see \S\ref{subsec:metrics}), with gains of around +25 points (76\% higher average score) compared to Normal RAG and Agentic RAG baselines.
    \item We analyze ablations, document type and task complexity effects, and cost--quality tradeoffs, showing that \textsc{SPD-RAG} attains over 85\% of full-context baseline quality at roughly 38\% of the API cost.
\end{itemize}

\section{Related Work}
\label{sec:related_work}

\subsection{LLM-Based Multi-Agent Systems}
\label{subsec:mas}

LLM-based multi-agent systems are increasingly used for complex task solving. \citet{guo2024large} provide a comprehensive survey of agent profiles, communication protocols, and collaboration strategies. Hierarchical architectures such as MegaAgent~\cite{wang2024megaagent}, which demonstrates autonomous cooperation among up to 590 agents via multi-level task decomposition, and AgentOrchestra~\cite{zhang2025agentorchestra}, which achieves state-of-the-art performance on GAIA using a central planning agent and specialized sub-agents, exemplify this trend.

Scaling laws for multi-agent systems offer quantitative guidance for architecture design. \citet{kim2025towards} evaluate 180 configurations across five architectures, finding that centralized coordination yields an 80.9\% improvement on parallelizable tasks, but that multi-agent overhead grows superlinearly. Their result that independent agents amplify errors by 17.2$\times$ compared to 4.4$\times$ for centralized systems directly motivates our coordinator-based design. The taxonomy of hierarchical multi-agent systems by \citet{moore2025taxonomy} further classifies design patterns along five axes: control hierarchy, information flow, role delegation, temporal layering, and communication structure.

\subsection{Long Document and Multi-Document QA}
\label{subsec:longdoc}

Long documents that exceed LLM context windows are often handled via divide-and-conquer strategies. LLM$\times$MapReduce~\cite{zhou2025llm} formalizes two key challenges: \textit{inter-chunk dependency} (one chunk depends on context from another) and \textit{inter-chunk conflict} (chunks provide contradictory information), and proposes structured information protocols and in-context confidence calibration to mitigate them. ToM~\cite{guo2025tom} extends this approach with a hierarchical DocTree representation for recursive reasoning.

For multi-document QA, LongAgent~\cite{zhao2024longagent} splits a 128K-token document into chunks assigned to member agents, with a leader agent orchestrating discussions and inter-member communication to reduce hallucinations. However, LongAgent targets a \textit{single} long document rather than \textit{multiple independent} documents that may conflict. DocAgent~\cite{sun2025docagent} uses a multi-agent framework that mimics human reading via a tree-structured outline and reviewer agent, while MDocAgent~\cite{han2025mdocagent} employs five specialized agents for multi-modal document understanding.

\subsection{Hierarchical and Recursive Summarization}
\label{subsec:hierarchical_summ}

Our hierarchical merging mechanism builds on tree-based retrieval and recursive summarization. RAPTOR~\cite{sarthi2024raptor} recursively embeds, clusters, and summarizes text chunks to construct a bottom-up tree with multiple abstraction levels; retrieving from the appropriate level yields a 20\% absolute accuracy gain on QuALITY. \citet{ou2025context} show that hierarchical merging can amplify hallucinations during recursive summarization and propose context-aware augmentation strategies that replace intermediate summaries with relevant input context, which we adopt in our merging protocol.

\subsection{Graph-Based Document Analysis}
\label{subsec:graphrag}

Graph-based document analysis is particularly effective for global sensemaking queries. GraphRAG~\cite{edge2024local} constructs a knowledge graph of entities and relations from source documents and applies hierarchical community detection (e.g., the Leiden algorithm~\cite{traag2019louvain}) to obtain multi-level communities with textual summaries. At query time, it retrieves relevant communities, elicits community-level partial answers, and aggregates them via a map--reduce--style procedure. A survey on LLM-empowered knowledge graph construction~\cite{bian2025llm} highlights how LLMs are reshaping the classical knowledge graph pipeline across three key stages: ontology engineering, knowledge extraction, and knowledge fusion. Our work is complementary: instead of inducing an explicit knowledge graph, \textsc{SPD-RAG} operates directly over agent-produced textual summaries and performs similarity-guided recursive synthesis across many documents.

\subsection{Benchmarks for Multi-Document Reasoning}
\label{subsec:benchmarks}

The Loong benchmark~\cite{wang2024leave} evaluates long-context LLMs on extended multi-document QA where every document is relevant ("Leave No Document Behind") across financial, legal, and academic domains in English and Chinese. It spans four task types with context lengths from 10K to over 250K tokens, and even strong frontier models achieve only modest Avg Scores and medium perfect-answer rates, underscoring the difficulty of reasoning over many long, all-relevant documents.

Complementary datasets probe other aspects of multi-document reasoning. MoNaCo~\cite{wolfson2025monaco} focuses on natural questions over dozens of Wikipedia pages, MEBench~\cite{lin2025mebench} on high-entity-density QA, FanOutQA~\cite{zhu2024fanoutqa} on multi-hop questions with long cross-document evidence chains, and HoloBench~\cite{maekawa2024holistic} on database-style aggregation over unstructured text. Taken together, these benchmarks show that scalable multi-document reasoning remains an open challenge, with Loong emphasizing the especially demanding all-relevant, long-context regime.

\section{Methodology}
\label{sec:methodology}

\begin{figure*}[t!]
    \centering
    \includegraphics[width=1.02\textwidth]{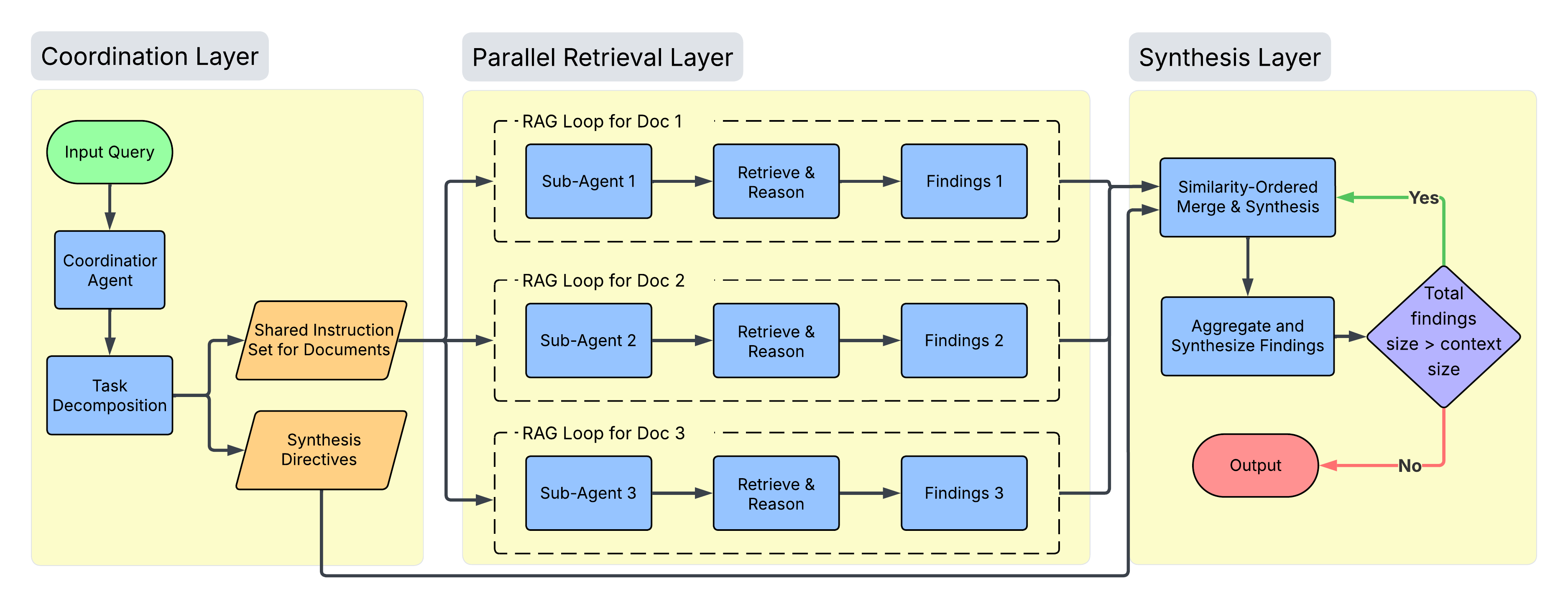}
    \caption{Overview of \textsc{SPD-RAG} Architecture.}
    \label{fig:architecture}
\end{figure*}

Figure~\ref{fig:architecture} presents an overview of the \textsc{SPD-RAG} architecture. The system processes a query through three layers: the Coordination Layer (\S\ref{subsec:coordination_layer}), the Parallel Retrieval Layer (\S\ref{subsec:retrieval_layer}), and the Synthesis Layer (\S\ref{subsec:synthesis_layer}).

\subsection{Coordination Layer}
\label{subsec:coordination_layer}

Given a query $q$ and corpus $\mathcal{D}$, the system first processes the query through a coordinator agent. The coordinator's role is to decompose the user query into a \textit{Shared Instruction Set for Documents} and \textit{Synthesis Directives}. It generates a \texttt{WriteTodos} structured object containing (a)~a list of \texttt{sub\_agent\_todos}: atomic, self-contained extraction tasks specifying exactly what fields, entities, or numeric values to extract (the shared instructions); and (b)~a \texttt{synthesis\_directive} (2--4 sentences) instructing the downstream Synthesis Layer on how to prioritize and structure the merged response.

\subsection{Parallel Retrieval Layer}
\label{subsec:retrieval_layer}

We assign a dedicated sub-agent $\alpha_i$ to each document $d_i$ in the corpus. Each sub-agent operates as an independent RAG loop over its assigned document, guided by the shared instruction set from the Coordination Layer.

\paragraph{Document-Scoped Retrieval and Reasoning.}
Each document $d_i$ serves as an isolated retrieval universe; the assigned sub-agent is strictly constrained to search tool calls within $d_i$, preventing cross-document distractor chunks from degrading local extraction. Each sub-agent receives the original query~$q$, its assigned document name, and the shared instruction set (todo list), then operates in an iterative retrieve-and-reason loop. On each turn, the agent either issues a \texttt{search} action (a focused query string) or a \texttt{finalize} action that terminates the loop and emits the document-grounded findings. The sub-agent prompt instructs the agent to attempt at least 2 focused searches before concluding a piece of information is absent, with a total cap of 5 search calls across all tasks. Each \texttt{search} action triggers dense vector retrieval over the sub-agent's dedicated per-document Qdrant index. This retrieval uses Cohere \texttt{embed-v4.0} embeddings (1536-dimensional, cosine similarity) to fetch the top $k=15$ most similar chunks. These chunks are then re-ranked by Cohere \texttt{rerank-v4.0-fast} to yield a final set of $\text{top\_n}=5$ chunks per search. The documents are split into chunks offline using LangChain's \texttt{RecursiveCharacterTextSplitter} in Markdown mode with a chunk size of 1000 tokens and 250-token overlap.

\paragraph{Findings Output.}
We denote the findings of each sub-agent as
\begin{equation}
    o_i = \alpha_i(q, d_i) = \langle s_i, r_i \rangle
\end{equation}
where $s_i$ is the natural language findings report from $d_i$ and $r_i \in [0,1]$ is a scalar relevance/confidence score. These outputs are passed to the Synthesis Layer.

\paragraph{Parallelism.}
All RAG loops for the documents are dispatched concurrently via LangGraph's \texttt{Send} API fan-out, which emits one \texttt{Send("document\_sub\_agent\_node", ...)} per document and executes all targets in parallel within LangGraph's async runtime.

\subsection{Synthesis Layer}
\label{subsec:synthesis_layer}

While modern LLMs feature extensive context windows, synthesizing hundreds of document reports can still exceed these limits or cause 'lost in the middle' degradation. To future-proof the architecture for massive corpora, the Synthesis Layer is designed as a dynamic map-reduce pipeline that aggregates findings recursively.
The Synthesis Layer receives the synthesis directives from the Coordination Layer and the findings $\{o_1, o_2, \ldots, o_n\}$ from the Parallel Retrieval Layer. It aggregates and synthesizes the findings recursively through a similarity-ordered process until the total findings size falls within the target context size to produce the final output.

\paragraph{Similarity-Ordered Merge \& Synthesis.}
In iteration $t$, we maintain a set of findings/summaries $\mathcal{S}^{(t)} = \{s^{(t)}_1, \ldots, s^{(t)}_n\}$. We embed each summary using Cohere \texttt{embed-v4.0} and compute a cosine-similarity matrix via \texttt{sklearn.metrics.pairwise.cosine\_similarity}. This is converted to a distance matrix $D_{ij} = 1 - \cos(s^{(t)}_i, s^{(t)}_j)$, which is passed to \texttt{AgglomerativeClustering} with \texttt{n\_clusters=1}, \texttt{linkage="average"} (UPGMA), producing a complete dendrogram over all current summaries. We traverse the merge tree bottom-up, greedily accumulating nodes into a batch as long as the total \texttt{tiktoken} count does not exceed the target budget of $B = 750{,}000$ tokens. Semantically similar summaries are grouped first, so each batch contains the most similar available summaries within the token limit.
    
\paragraph{Aggregate and Synthesize Findings.}
For each batch, an LLM is prompted with the concatenated findings, the synthesis directive, and the original query $q$ to produce an aggregated and synthesized summary. All batch synthesis calls within an iteration are issued concurrently via \texttt{asyncio.gather}. The merged summaries form the input for the next iteration $\mathcal{S}^{(t+1)}$. The same \textsc{Synthesis} prompt used at every level of the merging tree is applied even at the final level---there is no separate synthesis-only prompt.

\paragraph{Context Size Condition and Output.}
The layer evaluates whether the total findings size exceeds the context size threshold (i.e., $|\mathcal{S}^{(t+1)}| > 1$). If the condition is true, the process loops back to the similarity-ordered merge and synthesis step. If the condition is false (the total findings size is within the context size, leaving a single final summary $s^\star$), the process terminates and yields $s^\star$ as the final output. If $|\mathcal{S}^{(t+1)}| \geq |\mathcal{S}^{(t)}|$ (no-progress case), the entire current level is forced into a single batch to guarantee termination.

Algorithm~\ref{alg:merge} gives the pseudocode for the general recursive synthesis procedure.

\begin{algorithm}[t]
\caption{Recursive Synthesis}
\label{alg:merge}
\begin{algorithmic}[1]
\REQUIRE Findings $\mathcal{S} = \{s_1,\ldots,s_n\}$, query $q$, \\
\quad synthesis directive $d$, context size budget $B$
\ENSURE Final Output $s^\star$
\WHILE{$|\mathcal{S}| > 1$}
    \STATE $E \leftarrow \textsc{CohereEmbed}(\mathcal{S})$
    \STATE $D \leftarrow 1 - \textsc{CosineSimilarity}(E)$
    \STATE $\textit{tree} \leftarrow \textsc{AgglomerativeCluster}($ \\
    \qquad $D,\; \textit{linkage}=\text{``average''})$
    \STATE $\textit{batches} \leftarrow \textsc{GroupByTokens}($ \\
    \qquad $\mathcal{S},\; \textit{tree},\; B)$
    \IF{$|\textit{batches}| \geq |\mathcal{S}|$}
        \STATE $\textit{batches} \leftarrow [\mathcal{S}]$ 
    \ENDIF
    \STATE $\mathcal{S} \leftarrow \textsc{AsyncGather}($ \\
    \qquad $\textsc{Synthesize}(b, q, d)\ \forall b \in \textit{batches})$
\ENDWHILE
\RETURN $\mathcal{S}[0]$
\end{algorithmic}
\end{algorithm}

\paragraph{Behavior on Loong.}

In practice, with Gemini 2.5 Pro's 1M-token context window and a budget of $B = 750{,}000$ tokens, the combined sub-agent findings for all instances in our Loong evaluation set fit within a single batch, so the conditional check allowed output after one iteration. The recursive loop is therefore a design capability intended for much larger corpora, where total findings size genuinely exceeds the context size.

\section{Experimental Setup}
\label{sec:experiments}

\subsection{Dataset}
\label{subsec:dataset}

We evaluate \textsc{SPD-RAG} on the Loong benchmark ~\cite{wang2024leave}. Loong is an extended multi-document QA instances in both English and Chinese, with an average of 11 documents per test case spanning two real-world scenarios: financial reports and academic papers. Each test case is constructed so that every document is relevant to the final answer ("Leave No Document Behind", meaning a correct response requires synthesizing evidence from all provided documents), and is annotated under one of four task types: Spotlight Locating, Comparison, Clustering, or Chain of Reasoning, with context lengths ranging from 10K to beyond 250K tokens. We evaluate on the \textbf{English} and \textbf{Set 4 (200k-250k tokens)} portion of the benchmark only. Our evaluation set comprises 40 instances involving academic papers and 62 involving financial reports (102 instances total), with the four task types distributed as: Spotlight Locating (27), Comparison (15), Clustering (49), and Chain of Reasoning (11).

\subsection{Baselines}
\label{subsec:baselines}

We compare against the following systems, all backed by Gemini 2.5 Pro:

\begin{itemize}
    \item \textbf{Baseline (Full Context)}: All documents for each instance are concatenated and provided directly in the context window of Gemini 2.5 Pro, representing the oracle long-context upper bound.
    \item \textbf{Normal RAG}: Standard vector search top-$K$ retrieval over the full document corpus, followed by LLM reasoning over the retrieved chunks. 
    \item \textbf{Agentic RAG}: A single-agent ReAct-style RAG system (implemented via LangGraph/LangChain) that can issue multiple iterative retrieval calls over the full global corpus. It uses the same embedding and re-ranking pipeline as \textsc{SPD-RAG}, but lacks per-document specialization.
\end{itemize}

\subsection{Evaluation Metrics}
\label{subsec:metrics}

We report the following metrics, following the Loong evaluation protocol~\cite{wang2024leave}:
\begin{itemize}
    \item \textbf{Avg Score}: An LLM-judged score (0--100) measuring the degree to which the predicted answer accurately covers the gold answer, following the evaluation protocol of Loong \citep{wang2024leave}. \footnote{Because the original legacy GPT-4 model API is unavailable for new scaled evaluations, we adopt GPT-5 as the evaluator. Recent literature increasingly adopts GPT-5 as the standard for LLM-as-a-judge protocols due to its superior alignment with human grading \citep{wolfson2025monaco}. While our baseline numbers are re-computed under this GPT-5 judge for strict fairness, raw scores may not be perfectly 1-to-1 comparable with the originally published Loong leaderboard.}
    \item \textbf{Perfect Rate (PR\%)}: Percentage of queries where the system receives a perfect score  of 100.
    \item \textbf{Avg Token Usage}: Mean per-query total (input + output) token usage.
    \item \textbf{Avg Cost (USD)}: Mean per-query API cost, enabling cost-efficiency comparisons.
    \item \textbf{Avg Latency (s)}: Mean per-query wall-clock time.
\end{itemize}

\subsection{Implementation Details}
\label{subsec:implementation}

All systems use \textbf{Gemini 2.5 Pro} (temperature 0.0) as the backbone LLM for the coordinator, merging layer, and synthesis layer. \textsc{SPD-RAG} document sub-agents use \textbf{Gemini 2.5 Flash} for document-scoped retrieval, reducing per-document cost while preserving reasoning quality at the merging and synthesis stages. GPT-5 is used exclusively as the LLM judge for the evaluation of the Avg score and is not part of the pipeline itself.
Prompts for all systems, including baselines, were optimized for information extraction and synthesis to ensure fair comparison. Full prompt templates for all systems are provided in the code repository.

\paragraph{Indexing.} Each document is pre-indexed into a dedicated Qdrant vector collection using LangChain's \texttt{RecursiveCharacterTextSplitter} (Implementation details are provided in the Appendix.).

\paragraph{Retrieval per sub-agent.} Each \texttt{search} tool call performs dense vector retrieval over the document index, returning the top-$K$ ($k=15$) chunks using cosine similarity in Qdrant. These candidates are subsequently re-ranked with Cohere \texttt{rerank-v4.0-fast}, and the top $\text{top\_n}=5$ chunks are passed to the sub-agent LLM.

\paragraph{Merging layer.} The token budget per merge batch is $B = 750{,}000$ tokens (75\% of Gemini 2.5 Pro's 1M context cap). Agglomerative clustering uses \texttt{sklearn} with UPGMA linkage and precomputed cosine-distance matrix.
\paragraph{Parallelism.} Document agents run concurrently via LangGraph \texttt{Send} API fan-out. Batch synthesis calls within each merge iteration are parallelized with \texttt{asyncio.gather}. Synchronous Cohere API calls are offloaded to the thread pool via \texttt{asyncio.to\_thread}.

\paragraph{Safety limits.} Each document sub-agent's iterative retrieval loop is capped at 5 iterations; the Agentic RAG baseline caps at 10 iterations.

\paragraph{Cost tracking.} Per-query costs are computed from LangSmith run metadata (total token counts $\times$ per-token pricing.

\section{Results}
\label{sec:results}

\subsection{Main Results}
\label{subsec:main_results}

\begin{figure*}[t] 
    \centering
    \includegraphics[width=0.95\textwidth]{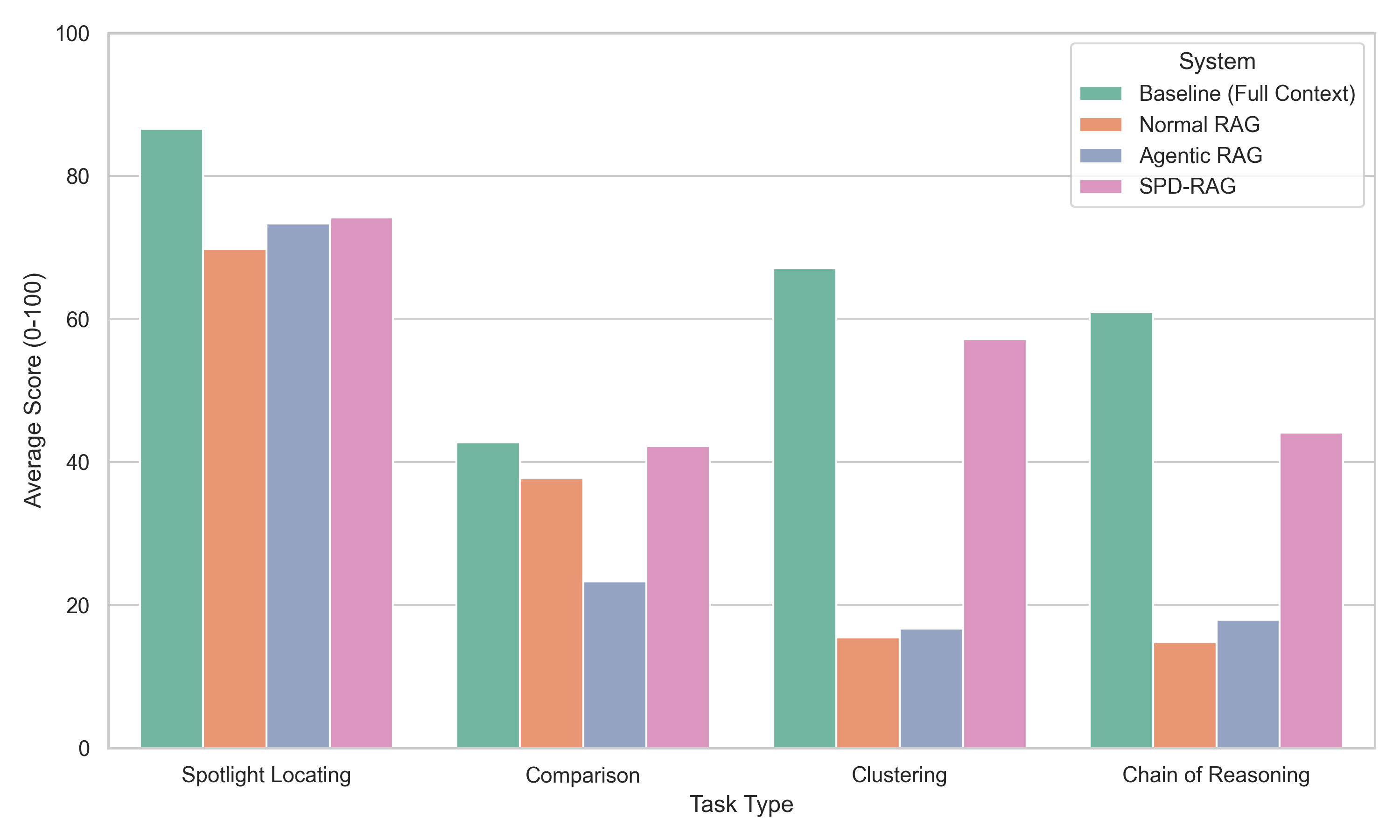}
    \caption{Comparison of Average Score across the four systems, broken down by task type (Spotlight Locating, Comparison, Clustering, and Chain of Reasoning).}
    \label{fig:main_results_chart}
\end{figure*}

Table~\ref{tab:main_results} reports the main results across all systems on our Loong evaluation set. Figure~\ref{fig:main_results_chart} illustrates these performance differences across the four specific task types.

\begin{table*}[t]
\centering
\small
\begin{tabular}{lcccccc}
\toprule
\textbf{System} & \textbf{Avg Score} & \textbf{PR (\%)} & \textbf{Avg Input Tok} & \textbf{Avg Total Tok} & \textbf{Avg Cost (\$)} & \textbf{Avg Latency (s)} \\
\midrule
\multicolumn{7}{l}{\textit{Oracle Long-Context Baseline}} \\
Baseline (Full Context) & \textbf{68.0} & \textbf{31.4} & 253,085 & 255,345 & 0.273 & 45.6 \\
\midrule
\multicolumn{7}{l}{\textit{RAG Baselines}} \\
Normal RAG              & 33.0          & 13.7           & \underline{\textbf{22,174}}  & \underline{\textbf{27,430}}  & \underline{\textbf{0.080}} & 42.6 \\
Agentic RAG             & 32.8          & 8.8            & 81,877  & 85,290  & 0.098              & \underline{\textbf{40.6}} \\
\midrule
\multicolumn{7}{l}{\textit{Our System}} \\
\textsc{SPD-RAG}        & \underline{58.1} & \underline{18.6} & 193,954 & 205,683 & 0.103 & 54.8 \\
\bottomrule
\end{tabular}
\caption{Main results on our Loong evaluation set (102 instances: 40 academic paper, 62 financial report). \textbf{Bold}: best overall. \underline{Underline}: best among RAG-based systems.}
\label{tab:main_results}
\end{table*}

\textsc{SPD-RAG} achieves an Avg Score of 58.1, substantially outperforming both Normal RAG (33.0) and Agentic RAG (32.8). This corresponds to roughly a 25-point absolute improvement (about 76\% relative gain) over standard RAG baselines. Furthermore, \textsc{SPD-RAG} more than doubles the Perfect Rate (PR) of Agentic RAG (18.6\% vs. 8.8\%), indicating that exhaustive document-level processing more frequently captures the complete set of required facts. Notably, the Agentic RAG baseline, despite consuming $\sim$3$\times$ more tokens than Normal RAG, does not yield score improvements, suggesting that uninstructed iterative retrieval without per-document specialization does not effectively address the coverage challenge in exhaustive multi-document settings. While the full-context baseline attains the highest quality with an Avg Score of 68.0, \textsc{SPD-RAG} reaches 85.4\% of this performance while consuming only 37.9\% of the API cost, demonstrating a highly favorable cost-quality trade-off. The full-context baseline provides a strong upper bound that benefits from all documents in one pass. Despite this, \textsc{SPD-RAG} narrows the gap to 9.9 points while costing less.

\subsection{Analysis by Task Type}
\label{subsec:complexity_analysis}

Table~\ref{tab:by_task} breaks down performance by Loong's four task types.

\begin{table*}[t]
\centering
\small
\begin{tabular}{lcccc}
\toprule
\textbf{System} & \textbf{Spotlight Loc.} & \textbf{Comparison} & \textbf{Clustering} & \textbf{Chain of Reasoning} \\
 & ($n=27$) & ($n=15$) & ($n=49$) & ($n=11$) \\
\midrule
Baseline (Full Context) & \textbf{86.6} / \textbf{55.6\%}  & \textbf{42.7} / \textbf{26.7\%}  & \textbf{67.1} / \textbf{14.3\%}  & \textbf{60.9} / \textbf{54.5\%} \\
Normal RAG              & 69.7 / 37.0\%        & 37.7 / \underline{26.7\%}        & 15.4 / 0.0\%                     & 14.8 / 0.0\% \\
Agentic RAG             & 73.4 / 22.2\%                    & 23.3 / 20.0\%                    & 16.7 / 0.0\%                     & 17.9 / 0.0\% \\
\textsc{SPD-RAG}        & \underline{74.2} / \underline{44.4}\%        & \underline{42.2} / 20.0\%        & \underline{57.2} / \underline{4.1\%}   & \underline{44.1} / \underline{18.2\%} \\
\bottomrule
\end{tabular}
\caption{Avg Score / PR(\%) by Loong task type. Each cell shows \textit{Avg Score} / \textit{PR\%}.}
\label{tab:by_task}
\end{table*}

The task-type breakdown reveals a clear pattern. For Spotlight Locating---which requires finding a single salient fact across documents---all RAG-based systems perform reasonably well (69--74), while the full-context baseline leads (86.6). The largest gaps between \textsc{SPD-RAG} and the RAG baselines appear in Clustering (+40.5 pts over Normal RAG) and Chain of Reasoning (+26.2 pts over Agentic RAG), tasks that require aggregating or reasoning over evidence from \textit{many} documents simultaneously. On the other hand, it nearly matches the full-context baseline (42.2 vs.\ 42.7) on Comparison tasks.

\subsection{Analysis by Document Domain}
\label{subsec:domain_analysis}

\begin{table}[t]
\centering
\small
\begin{tabular}{lcc}
\toprule
\textbf{System} & \textbf{Paper} ($n=40$) & \textbf{Financial} ($n=62$) \\
\midrule
Baseline        & \textbf{78.8} / 30.0\% & \textbf{61.0} / 32.3\% \\
Normal RAG      & 15.2 / 0.0\%           & 44.5 / 22.6\% \\
Agentic RAG     & 16.8 / 0.0\%           & 43.1 / 14.5\% \\
\textsc{SPD-RAG}& \underline{60.0} / 7.5\%  & \underline{56.9} / \underline{25.8\%} \\
\bottomrule
\end{tabular}
\caption{Avg Score / PR(\%) by document domain.}
\label{tab:by_domain}
\end{table}

\begin{figure}[htbp]
    \centering
    \includegraphics[width=0.9\linewidth]{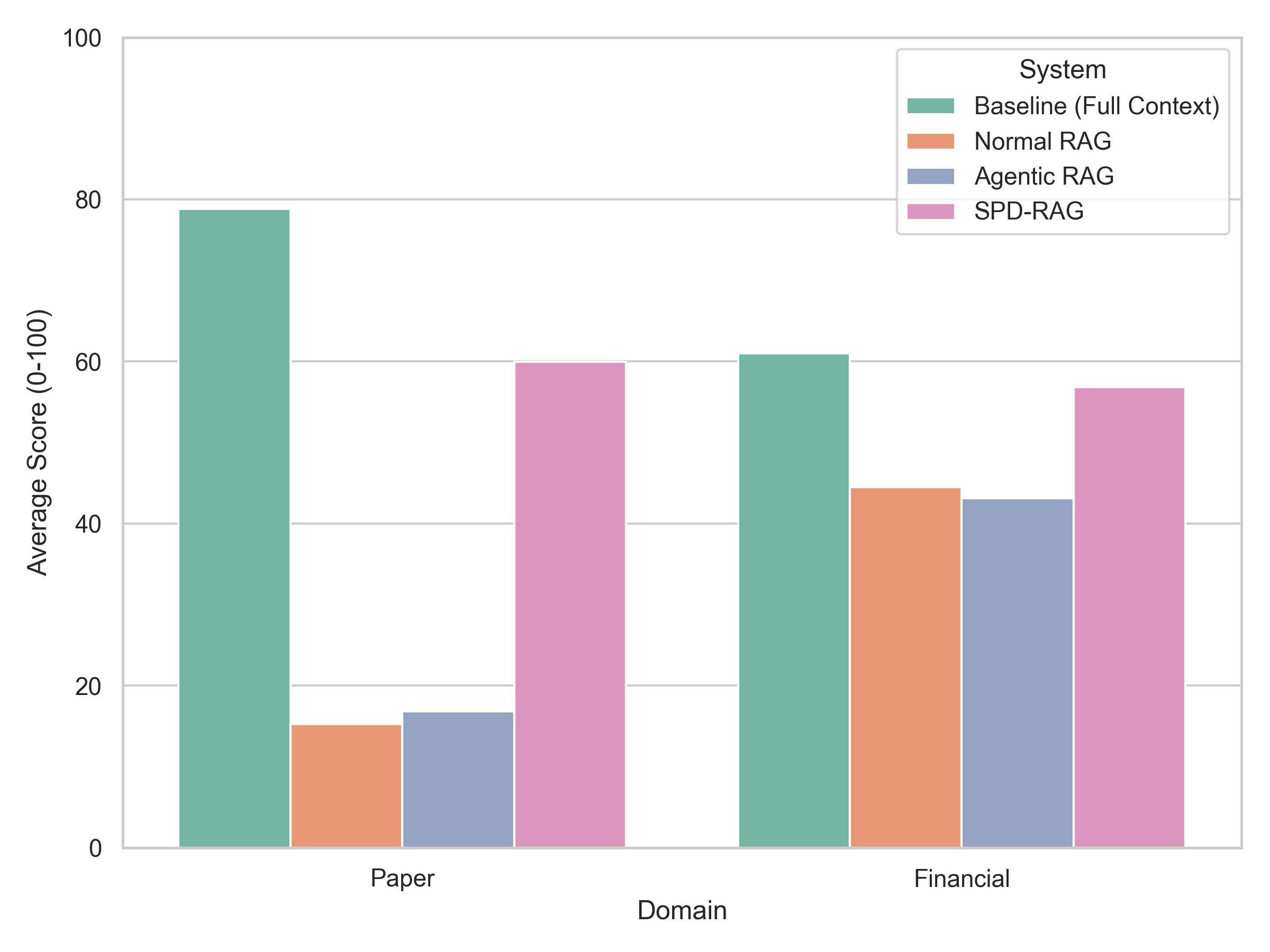}
    \caption{Average score by document domain comparing the Baseline (Full Context), Normal RAG, Agentic RAG, and SPD-RAG systems.}
    \label{fig:domain_chart}
\end{figure}

As shown in~\autoref{tab:by_domain} and ~\autoref{fig:domain_chart} the domain breakdown exposes a striking failure mode of standard RAG: both Normal RAG and Agentic RAG achieve 0\% PR on academic paper instances (Avg Scores of 15.2 and 16.8, respectively). Academic paper instances in Loong tend to involve long, technical documents with distributed evidence, making top-$K$ retrieval especially susceptible to coverage failures. \textsc{SPD-RAG} dramatically recovers on this domain (60.0 Avg Score), closing most of the gap with the full-context baseline (78.8). On financial report instances---which tend to be shorter and more structured---all RAG-based systems perform better, and the gap between \textsc{SPD-RAG} (56.9) and the baseline (61.0) is smaller.

\subsection{Cost--Quality Tradeoff}
\label{subsec:cost_quality}

\begin{table}[t]
\centering
\small
\begin{tabular}{lccc}
\toprule
\textbf{System} & \textbf{Avg Score} & \textbf{Avg Cost (\$)} & \textbf{Score / \$} \\
\midrule
Baseline        & \textbf{68.0} & 0.273 & 249.1 \\
Normal RAG      & 33.0 & \textbf{0.080} & 412.5 \\
Agentic RAG     & 32.8 & 0.098 & 334.7 \\
\textsc{SPD-RAG}& 58.1 & 0.103 & \textbf{564.1} \\
\bottomrule
\end{tabular}
\caption{Cost--quality tradeoff.}
\label{tab:cost_quality}
\end{table}

\begin{figure}[htbp]
    \centering
    \includegraphics[width=0.9\linewidth]{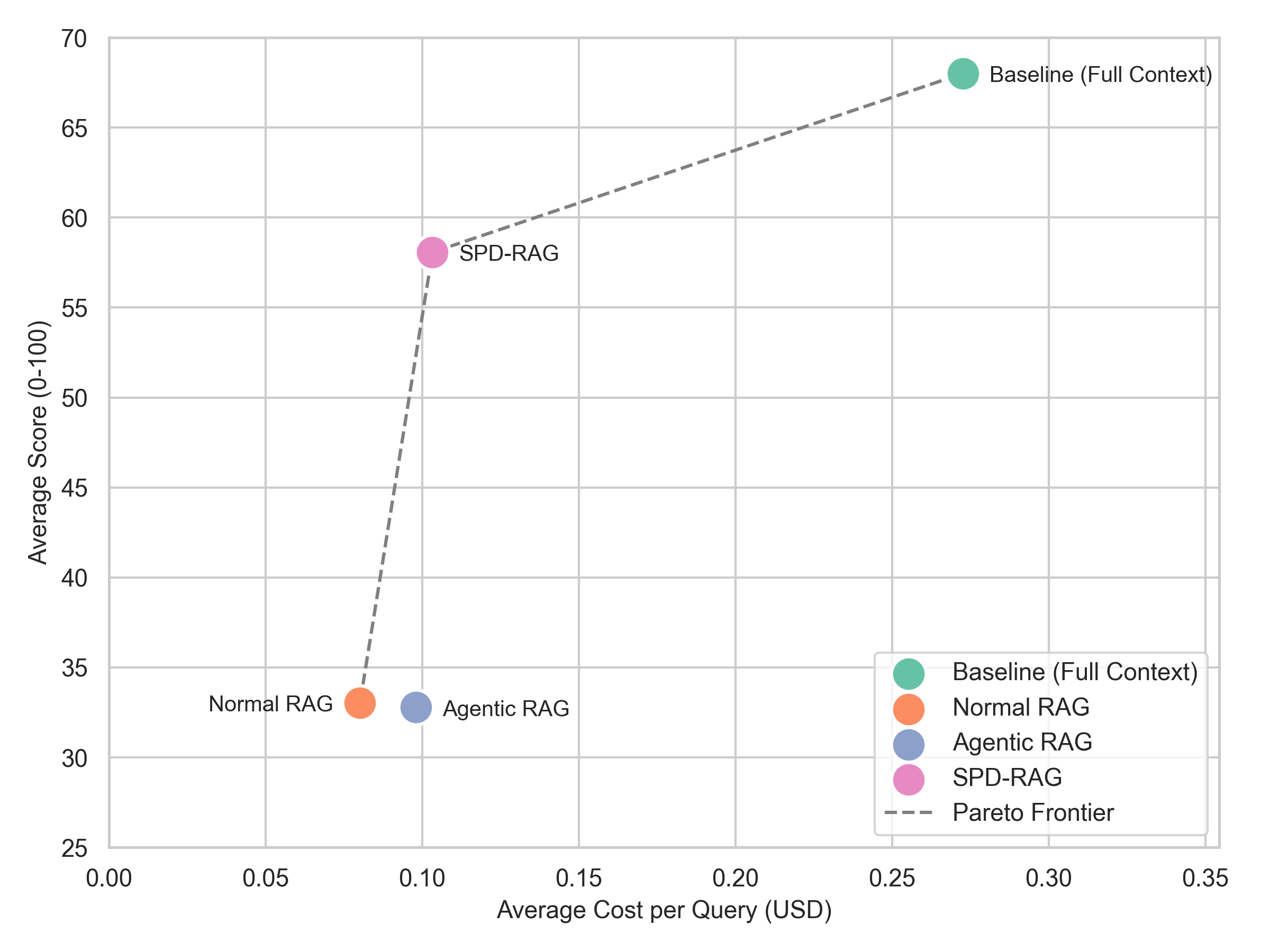}
    \caption{Cost--Quality tradeoff. The scatter plot illustrates the Pareto frontier for multi-document QA.}
    \label{fig:cost_quality_chart}
\end{figure}

\textsc{SPD-RAG} outperforms the RAG baselines in the cost--quality trade-off. It achieves a \textbf{76\% higher average score} than Normal RAG (58.1 vs.\ 33.0) while increasing the per-query cost by only \$0.023 (\autoref{tab:cost_quality}). In contrast, Agentic RAG is \textbf{Pareto-dominated}, as it incurs higher cost than Normal RAG (\$0.098 vs.\ \$0.080) while delivering nearly identical quality (32.8 vs.\ 33.0). Compared to the full-context baseline, which achieves an Avg Score of \textbf{68.0}, \textsc{SPD-RAG} reaches \textbf{85.4\% of the quality at only 37.9\% of the cost}, resulting in a \textbf{2.25$\times$ improvement in cost--quality efficiency}.
Crucially, the architecture of \textsc{SPD-RAG} enables this cost efficiency: by constraining retrieval to isolated, single-document spaces, we can reliably offload the iterative reasoning to a cheaper model (Gemini 2.5 Flash). In contrast, global iterative baselines require the reasoning capacity of a frontier model (Gemini 2.5 Pro) to navigate the fully concatenated corpus, driving up costs.

\section{Discussion}
\label{sec:discussion}

\paragraph{Where SPD-RAG excels.}
The task-type results in~\autoref{tab:by_task} confirm our core hypothesis: architectural decomposition along the document axis yields the largest gains precisely for tasks that require exhaustive multi-document synthesis. Clustering tasks (+40.5 over Normal RAG) and Chain of Reasoning tasks (+26.2 over Agentic RAG) both demand that the system form a coherent answer from evidence distributed across all relevant documents---exactly the scenario for which per-document agents are designed. For Loong's Comparison questions, the answer is primarily determined by reading out a small set of comparable numeric attributes from each report and aggregating them, so once document agents have extracted those fields, the synthesizer is effectively operating over a structured table, which helps explain why \textsc{SPD-RAG}'s score on Comparison tasks (42.2) nearly matches the full-context baseline (42.7).

\paragraph{Remaining gap with the full-context baseline.}
Despite strong gains over RAG baselines, \textsc{SPD-RAG} lags the full-context baseline by 9.9 Avg Score points overall. Three factors likely contribute:~(1)~Gemini 2.5 Flash, used for document agents due to its cost efficiency, may provide weaker agentic reasoning capabilities compared to larger models.~(2)~The coordinator's sub-task generation may under-specify queries for highly technical academic content, leading to incomplete information extraction at the document-agent level.~(3)~The maximum context length observed in Loong queries was approximately 250k tokens---only about 25\% of the 1M-token context window of Gemini 2.5 Pro---potentially limiting the pressure on the full-context baseline. Since \textsc{SPD-RAG} is designed to operate on substantially larger contexts, its advantages may become more pronounced under more extreme long-context conditions.

\paragraph{RAG baseline failure on papers.}
The 0\% PR rate and 15.2-16.8\% Avg Score of both Normal RAG and Agentic RAG on academic paper instances highlights a systematic failure of top-$K$ retrieval under dense, distributed evidence. Academic papers in Loong often require synthesizing facts spread across abstracts, methods, results, and appendices of multiple documents---a regime where exhaustive per-document coverage proves highly effective. The recovery by \textsc{SPD-RAG} to 60.0 Avg Score / 7.5\% PR in this domain demonstrates the practical value of per-document specialization.

\paragraph{Latency overhead.}
\textsc{SPD-RAG} incurs a modest latency overhead (54.8 s vs.\ 40.6--45.6 s for the baselines), primarily due to its multi-agent architecture. Although document agents run in parallel, end-to-end latency is still gated by the layered pipeline, which requires at least three sequential LLM calls (coordination, per-document retrieval, and synthesis) rather than a single pass.

\begin{figure}[htbp]
    \centering
    \includegraphics[width=0.85\linewidth]{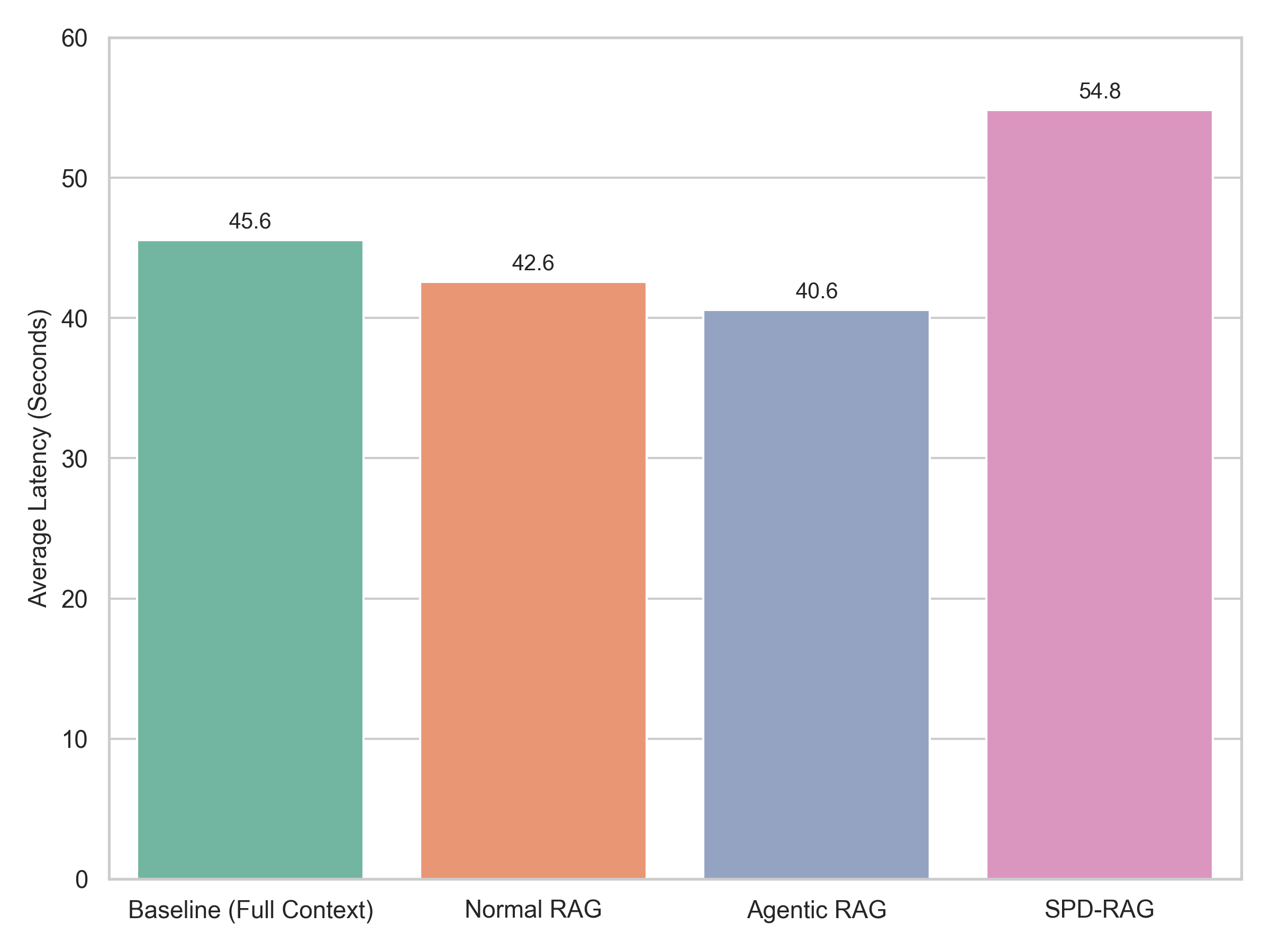}
    \caption{Average per-query latency across systems.}
    \label{fig:latency_chart}
\end{figure}

\section{Limitations and Future Work}

\textsc{SPD-RAG} incurs higher per-query LLM call counts than single-pass systems---one call per document agent plus a final synthesis step---which increases total token usage compared to standard RAG baselines, though it remains at roughly one-third the cost of the full-context baseline (\$0.10 vs.\ \$0.27 per query) due to the use of Gemini 2.5 Flash for document sub-agents. System quality depends on the coordinator's ability to generate comprehensive sub-tasks: under-specified tasks lead to incomplete document-level extractions, as observed on technical academic paper instances. Our evaluation is limited to two domains (academic papers and financial reports) from the Loong benchmark; legal, medical, or enterprise document corpora may have different structural characteristics that alter relative performance.

The synthesis layer of \textsc{SPD-RAG} is designed to handle corpora where the combined sub-agent outputs exceed the model's context window, triggering multi-round similarity-based clustering and recursive synthesis. In our Loong experiments, however, Gemini 2.5 Pro's 1M-token context was sufficient to process all sub-agent summaries in a single merge step, so this recursive path was never activated. As a result, the scalability properties of the recursive synthesis pipeline---its impact on answer quality, redundancy reduction, and cost as document counts grow into the hundreds or thousands---remain empirically unvalidated. Evaluating \textsc{SPD-RAG} on much larger corpora, and ideally on a new benchmark specifically targeting hundreds to thousands of documents per query instance, is an important direction for future work that would more directly test the per-document agent design and the intended large-database regime.

\section{Conclusion}
\label{sec:conclusion}

We presented \textsc{SPD-RAG}, a hierarchical multi-agent framework designed for exhaustive multi-document question answering. By assigning a dedicated agentic RAG module to each document and merging their findings through a similarity-ordered, token-bounded tree, our system effectively mitigates the limitations of both traditional top-$K$ retrieval (incomplete coverage) and isolated long-context processing (degraded reasoning at scale). Experiments on the challenging Loong benchmark demonstrate that \textsc{SPD-RAG} achieves an average score of 58.1, outperforming Normal RAG (33.0) and Agentic RAG (32.8) by approximately 25 absolute points (76\% higher average score), while operating at only 37.9\% of the API cost of the oracle full-context baseline (68.0). These performance gains are most pronounced in tasks requiring deep cross-document synthesis---such as Clustering (+40.5 points) and Chain of Reasoning (+26.2 points)---as well as on dense academic papers where standard RAG methods entirely fail.

Ultimately, our findings demonstrate that for complex information-seeking queries over large corpora, \textit{how} information is processed is crucial. Specifically, ensuring that each document receives exhaustive, dedicated agentic attention proves to be a more effective, cost-efficient, and scalable strategy than simply expanding \textit{how much} raw context a single model can consume in one pass.

\section{Ethics Statement}

\textsc{SPD-RAG} is a retrieval-augmented generation framework intended to improve access to information distributed across large document collections. The system relies on commercial LLM APIs (Gemini 2.5, Cohere), and as with any LLM-based pipeline, outputs may contain factual errors and should not be treated as authoritative without human verification. The increased number of API calls per query relative to single-pass systems results in higher energy consumption, a cost that should be weighed against the accuracy gains in deployment decisions. Our benchmark evaluations are conducted on publicly available datasets and involve no human subjects or personal data. During the preparation of this manuscript, AI-assisted writing tools were used to help refine language and improve clarity. All research design, experiments, analysis, and final decisions regarding the content were conducted and verified by the authors.

\bibliography{latex/custom}

\appendix

\section{Prompts}
\label{app:prompts}

All prompts are defined in \texttt{backend/core/prompts.py}.

\subsection*{A.1 Coordinator Prompt (\texttt{LEAD\_RESEARCHER\_PROMPT})}

\begin{quote}\small\ttfamily
You are a lead researcher coordinating a RAG-based analysis to answer a user query.

Pipeline Architecture: \\
1. YOU (now) --- Decompose the query into extraction tasks and write a synthesis directive to guide the downstream synthesizer. \\
2. Sub-agents (next) --- Each document is assigned one worker that runs your tasks independently. Workers search their document and report raw findings. They run in parallel and cannot see each other's results. \\
3. Synthesizer (last) --- A downstream agent receives ALL worker findings and merges them into a single coherent response. It follows your \texttt{synthesis\_directive} to decide what to prioritize and how to structure the output.

You CANNOT see the names, types, or content of the documents.

Your ONLY job in this step:
\begin{itemize}
\item Produce a list of \texttt{subagent\_todos} (as structured output): precise extraction tasks that will be executed independently against EACH document by the sub-agents.
\item Produce a \texttt{synthesis\_directive}: a concise instruction (2--4 sentences) for the synthesizer.
\end{itemize}

Todo-writing rules:
\begin{itemize}
\item Decompose the user query into concrete information requirements.
\item Each todo must be self-contained and unambiguous.
\item Prefer atomic tasks over broad tasks.
\item Include coverage for: definitions, numeric values, constraints, edge cases, error modes, and any explicit recommendations required by the user query.
\item Design a robust extraction list that works for ANY document in the set.
\end{itemize}

Important Constraints:
\begin{itemize}
\item Do not assume any document contains the answer. Write todos that can be answered with either ``Found'' or ``Not found in this document.''
\item Tell the worker WHAT to extract, not HOW to extract it.
\item Do not synthesize, summarize, or attempt to answer the user query yourself.
\end{itemize}
\end{quote}

\subsection*{A.2 Document Sub-Agent Prompt (\texttt{RESEARCH\_SYSTEM\_PROMPT})}

\begin{quote}\small\ttfamily
You are a sub-agent investigating a single document: \texttt{\{file\_name\}}.

You operate inside an iterative retrieval loop. On each turn you output exactly one structured action:

\begin{itemize}
\item \texttt{action="search"}: Issue a focused query to retrieve information from the document. Set \texttt{query} to a specific, targeted search string. Set \texttt{reasoning} to why this query is needed.
\item \texttt{action="finalize"}: You have gathered sufficient evidence to address ALL assigned tasks. Set \texttt{findings} to your complete extracted report. Set \texttt{reasoning} to a brief summary of what you found.
\end{itemize}

Investigation principles:
\begin{enumerate}
\item Start by identifying the key facts required for each task.
\item Issue ONE focused query per turn --- prefer specific terms over broad phrases.
\item Expand to synonyms or paraphrases if initial results are incomplete.
\item Issue separate queries for different aspects of a task.
\item Keep searching until every task is covered with concrete evidence or confirmed absent.
\item Do NOT attempt to answer the user query directly --- extract raw facts only.
\item Do NOT finalize a task as 'Not found' after only a single focused search. At least attempt 2 focused searches before concluding information is absent.
\item Do NOT focus on very specific terms, but rather on the general context of the task. Always try to find the most relevant information.
\item You must use AT MOST 5 searches to answer a task. DO NOT OVERUSE THE SEARCH TOOL.
\end{enumerate}

Finalize report format: For each assigned task, output either ``Found: [exact answer with supporting evidence]'' or ``Not found in this document.'' Always report exact numbers, names, and dates. Never approximate or infer beyond the document.
\end{quote}

\subsection*{A.3 Merge / Synthesis Prompt (\texttt{SYNTHESIS\_PROMPT})}

This prompt is used uniformly at \textit{every} level of the recursive merging tree, including the final synthesis step. There is no separate final-answer prompt.

\begin{quote}\small\ttfamily
You are a research synthesizer. Merge the following sub-agent findings into one compact, information-dense response for the user.

Query: \texttt{\{query \}}

Synthesis Directive: \\
\texttt{\{synthesis\_directive\}}

Findings Batch: \texttt{\{findings \}}

Rules:
\begin{itemize}
\item Follow the synthesis directive above --- it defines your main goal, priorities, and output structure.
\item Keep only information that directly helps answer the query. Discard tangential content.
\item Preserve exact numbers, names, dates, and caveats.
\item Remove redundancy. If the same fact appears multiple times, keep it once.
\item Do NOT invent or infer facts not present in the findings.
\item Same language as the findings (default English if mixed).
\end{itemize}
\end{quote}

\section{Implementation Details}
\label{app:implementation}

\paragraph{Models.} The coordinator, the merging layer, and the synthesis layer use \texttt{gemini-2.5-pro} (temperature 0.0). Document sub-agents use \texttt{gemini-2.5-flash} (temperature 0.0) for document-scoped retrieval. The LLM judge (GPT-5) is separate and is used only for the evaluation of the Avg score. Model constants are defined in \texttt{backend/shared/constants.py}.

\paragraph{Hyperparameters.}
\begin{center}
\small
\begin{tabular}{ll}
\toprule
\textbf{Parameter} & \textbf{Value} \\
\midrule
Chunk size (tokens) & 1000 \\
Chunk overlap (tokens) & 250 \\
Tokenizer & \texttt{tiktoken cl100k\_base} \\
Embedding model & Cohere \texttt{embed-v4.0} \\
Embedding dimension & 1536 \\
Similarity metric & Cosine \\
Merge token budget $B$ & 750,000 tokens \\
Dense retrieval top-$K$ & 15 \\
Reranker & Cohere \texttt{rerank-v4.0-fast} \\
Reranker top-$N$ & 5 \\
Sub-agent iteration safety limit & 5 \\
Clustering algorithm & Agglomerative (UPGMA) \\
Clustering metric & Precomputed cosine distance \\
\bottomrule
\end{tabular}
\end{center}

\paragraph{Cost measurement.} Per-query costs are pulled from LangSmith run metadata (\texttt{extract\_langsmith\_info.py}) using model-specific per-token pricing.

\end{document}